\documentclass[10pt, conference]{IEEEtran}

\usepackage{graphicx}
\usepackage{amsmath}
\usepackage{amssymb}
\usepackage{hhline}
\usepackage{array}

\usepackage{multirow}
\usepackage{tabularx}

\usepackage{subfigure}
\usepackage{diagbox}

\usepackage{xcolor}

\ifCLASSINFOpdf
\else
\fi

\DeclareMathOperator*{\argmaxA}{arg\,max} 
\DeclareMathOperator*{\argminA}{arg\,min} 
\begin{document}


%

\title{An AutoML-based Approach to \\Multimodal Image Sentiment Analysis}

\newcommand\blfootnote[1]{%
  \begingroup
  \renewcommand\thefootnote{}\footnote{#1}%
  \addtocounter{footnote}{-1}%
  \endgroup
}

\author{\IEEEauthorblockN{ Vasco Lopes\IEEEauthorrefmark{1}, António Gaspar\IEEEauthorrefmark{1},  Lu{\'{\i}}s A. Alexandre\IEEEauthorrefmark{1}, João Cordeiro\IEEEauthorrefmark{2,3}}
\IEEEauthorblockA{\IEEEauthorrefmark{1}NOVA LINCS, Universidade da Beira Interior}
\IEEEauthorblockA{\IEEEauthorrefmark{2}LIAAD, INESC TEC – Institute for Systems and Computer Engineering, Technology and Science}
\IEEEauthorblockA{\IEEEauthorrefmark{3}HULTIG – Centre of Human Language Technology and Bioinformatics}
\{vasco.lopes, antonio.pedro.gaspar, luis.alexandre, jpcc\}@ubi.pt
}


\newcommand{\ver}[1]{{\color{red}!!! #1 !!!}}  

\maketitle


\begin{abstract}
    Sentiment analysis is a research topic focused on analysing data to extract information related to the sentiment that it causes. Applications of sentiment analysis are wide, ranging from recommendation systems, and marketing to customer satisfaction. Recent approaches evaluate textual content using Machine Learning techniques that are trained over large corpora. However, as social media grown, other data types emerged in large quantities, such as images. Sentiment analysis in images has shown to be a valuable complement to textual data since it enables the inference of the underlying message polarity by creating context and connections. Multimodal sentiment analysis approaches intend to leverage information of both textual and image content to perform an evaluation. Despite recent advances, current solutions still flounder in combining both image and textual information to classify social media data, mainly due to subjectivity, inter-class homogeneity and fusion data differences. In this paper, we propose a method that combines both textual and image individual sentiment analysis into a final fused classification based on AutoML, that performs a random search to find the best model. Our method achieved state-of-the-art performance in the B-T4SA dataset, with 95.19\% accuracy.
    
\end{abstract}


%
\IEEEpeerreviewmaketitle

\section{Introduction}
    Sentiment analysis is an ever-growing research topic, where the focus is to analyze the underlying sentiment of a given source of data, based on its subjectivity and context. Sentiment analysis is mostly performed using textual data, where the goal is to, based on a sentence or a text, determine the author's message polarity. The classification is generally binary - either negative or positive, or $n$-class classification, wherein, the most used one is a 3-class classification - negative, neutral and positive, using either machine-learning approaches, where a classifier is trained using a labelled corpus, or using lexicon-based approaches, where the textual information is classified based on its semantics or by using statistical approaches \cite{MEDHAT20141093, aggarwal2012survey}. Applications of sentiment analysis can be seen in many contexts, such as brand awareness \cite{jansen2009twitter}, political voting intentions \cite{MEDHAT20141093}, customer satisfaction \cite{grabner2012classification, shayaa2018sentiment} and in disaster relief \cite{beigi2016overview}.
    
    The proliferation of social media opened doors for massive collection of data for sentiment analysis \cite{10.1145/2938640}. These are important channels of human communication, allowing for instantaneous spread of information. Twitter emerged as one focal point to capture and analyze data, in which users express their opinions, feelings and thoughts regarding entities or events. Detecting sentiments in Twitter differs from detecting sentiments in conventional text such as blogs and forums, due to the reduced size of the textual data, and because of the information context, addition of symbols in the form of emojis and irony and subjectivity. However, social media networks, such as Twitter, also provide the opportunity to congregate textual data with more information, usually in the form of images, videos or audio. The coupling of multiple data sources, allows the development of multimodal classification, in which a method leverages more than one type of data to perform classification \cite{soleymani2017survey}. This is significantly harder in the context of sentiment analysis, as extracting sentiments solely from textual information is easier than combining information from text and, for example, images. Even though a multimodal approach can improve the performance when compared to a sole text-approach \cite{majumder2018multimodal}, this is a challenging task, especially when using data acquired from social media, as the different data types are sparse and can have different contexts, present irony, different intentions and their combined evaluation is not trivial.

    In this paper, we propose a novel multimodal sentiment analysis method, that uses both textual data and images from social media to perform 3-class classification regarding polarity. The proposed method consists of initial individual classification of the textual and image components, and then, based on Automated Machine Learning (AutoML), fuse both classifications into a final one. To perform the individual classifications, we leverage the power of deep neural networks. For this, we evaluated the performance of multiple networks and then selected the best for both the text and the image part. Then, to build the fusing method, we used AutoML to perform a random search to determine the best model to perform the final classification. We evaluated the proposed method in the task of multimodal sentiment analysis using a dataset containing over 470 thousand tweets, where each tweet is composed of both textual and image content.


    \begin{figure*}[!t]
        \centering
        \includegraphics[width=1\textwidth]{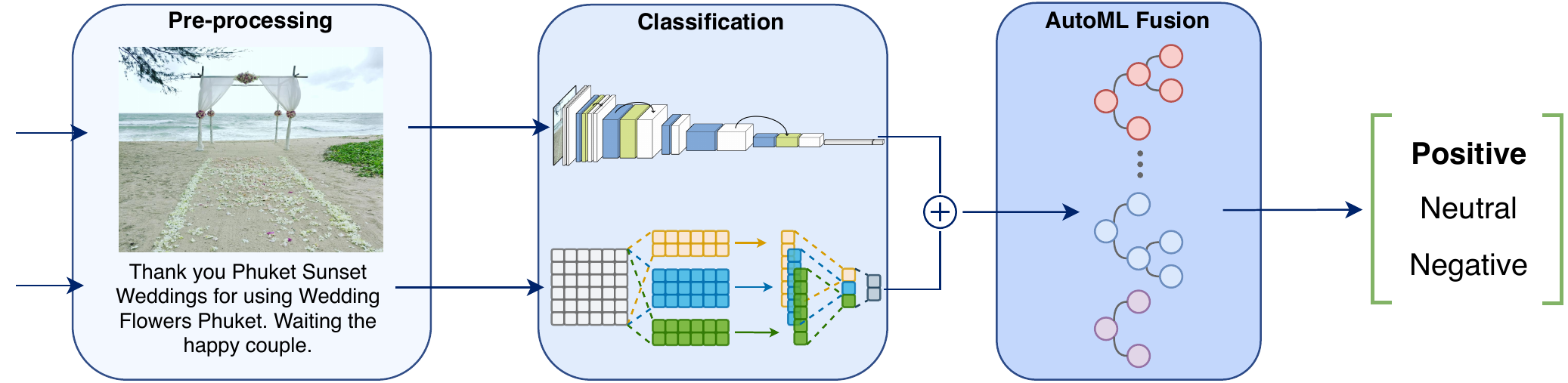}
        \caption{Proposed multimodal architecture. The first container represents the pre-processing component, that receives an image and associated text, and pre-processes it to remove noise and non-important data. The second container shows both classification components, where the image and the text are classified individually using CNNs. The third container receives the concatenation of the individual classifications, and performs a final classification using the optimal model searched - represented by a Gradient Boosting Machine (GBM) in the image.}
        \label{fig:architecture2}
    \end{figure*}

  The contributions of this paper, can be summarized as follows: 1) we conduct a comparison regarding the performance of sentiment analysis in textual data from Twitter; 2) we compare different state-of-the-art deep learning models in image sentiment analysis, and 3), we propose a novel fusion method that combines the individual classifications into a final one, by levering an optimal model generated with AutoML mechanisms, which resulted in state-of-the-art accuracy on the B-T4SA dataset.
  
  The remainder of this work is organized as follows. Section~\ref{related_work}, contextualizes the related work in multimodal sentiment analysis, and the use and application of AutoML methods. Section~\ref{proposed_method}, details the proposed method. Presenting the architecture of our proposal, including a detailed description of the text and image analysis methods, as well as the AutoML component. In Section~\ref{experiments}, we introduce the datasets used, a description of the conducted experiments, and discuss the results. Finally, Section~\ref{conclusions} presents a conclusion.


\section{Related Work}
\label{related_work}

    \subsection{Multimodal Sentiment Analysis}
    Even though the vast majority of the sentiment analysis proposals focus on single model sentiment analysis, mainly using textual information \cite{MEDHAT20141093, zhang2018deep}, there are interesting proposals that try to fuse more than one source of information to perform multimodal sentiment analysis. The method proposed in \cite{wollmer2013youtube}, combines features from audio and video, and fuses them with text features to estimate the sentiment of youtube movie reviews. In \cite{zadeh2017tensor}, the authors propose Tensor Fusion Network, which is a network capable of fusing features extracted from different sources of data, into a single tensor, allowing sentiment analysis both in separate and conjoined. Taking a different approach \cite{majumder2018multimodal}, performs multimodal sentiment analysis by conducting hierarchical fusion of the different features by first fusing them into pairs, and then combining them into one. In \cite{DBLP:conf/aaai/TruongL19}, the authors proposed Visual Aspect Attention Network, in which the goal is to use images as attention mechanisms to aid in detecting important sentences in documents. To perform such operation, images are analyzed using a CNN, and the output is used as weights in a word encoder. In~\cite{AGasparIdeal}, the authors propose a multimodal method that performs individual analysis of both the image and the text, and then performs a weighted average over the individual predictions to perform a final classification. However, while producing the individual predictions, the authors also introduce \textit{Image Content Analysis}, a second method to classify the images, which is based on detecting the most predominant object on the image and classifying the image based on the probability of that object appearing in a given class in the training set.

    \subsection{AutoML}
    AutoML \cite{automlbook}, focuses on developing approaches that provide efficient methods to design machine learning workflows without extensive need for human intervention or optimization processes \cite{he2020automl}. There have been proposals for solving the optimization problem of designing machine learning workflows using random search \cite{bergstra2012random}, evolutionary strategies \cite{liang2019evolutionary}, bayesian optimization \cite{kotthoff2017auto} and reinforcement learning \cite{he2018amc}. In \cite{kocbek2019automated}, the authors extended Auto-Weka in order to use AutoML to design methods to detect railway track defects. In \cite{lopes2020auto}, AutoML is leveraged to improve the classification component of CNNs. This method was built by training CNNs until convergence and then partially removing their classification component and replacing it by a searched model, resulting in performance improvements both in accuracy and inference time. Finally, in \cite{agrapetidou2020automl}, AutoML is used to forecasting bank failures by performing automated feature extraction.
    
    Our work has similarities with Neural Architecture Search (NAS) \cite{elsken2019neural}, in the sense that the focus is in building the classifier and not the entire process of a machine learning workflow, but the main difference is that we do not limit our search to neural networks, rather, we allow more machine learning models to be searched.

    Some of the differentiating points of our proposal are the fact that it performs a simple and efficient fusion, based on the individual analysis of the text and image components, and that it takes advantage of AutoML for finding the final classifier that works on top of the fused features.

\section{Proposed Method}
    \label{proposed_method}

\subsection{Architecture}
\label{Architecture}
    The proposed method is composed of three stages: pre-processing, individual classifications, and the fusion stage. The entire architecture of the method can be seen in Fig.~\ref{fig:architecture2}. For both individual classification components, we implemented several state-of-the-art methods to perform a comparison and select the most performant one on the validation set, to be integrated into the proposed architecture.
  
    In the following sections, we present the implementation details of both the image (\ref{image_analysis}) and text classification (\ref{text_analysis}), as well as the implementation details for the fusion mechanism (\ref{fusion:automl}). Moreover, in each section, we further explain the pre-processing components of the image and the text.

\subsection{Image Sentiment Analysis}
\label{image_analysis}
    To classify the post's polarity using image data, we explored the use of state-of-the-art CNNs that perform feature extraction and classification over the inner representations created. For this, we focused on using two architectures that are known to do well in different image analysis tasks: ResNet and DenseNet. The remainder of this section details the CNNs used, as well as the pre-processing steps for cleaning the input images.

    \subsubsection{Preprocessing} \hfill
    \label{img_pre} 
    
    We first apply a resize operation, changing the image size to $224*224$, to uniformize the dataset, as images from the Twitter dataset have different dimensions. Then, each image is normalized using the mean and standard deviation of each channel in the whole dataset. This is applied by subtracting the mean and dividing by the standard deviation in each channel of each image: $img_i = (o_i - \mu_i)/\sigma_i, \quad i = 1,..,c$, where $i$ represents the channel, $o$ the original image, $\mu$ is the mean of the dataset and $\sigma$ is the standard deviation of the dataset.

    \subsubsection{Models}\hfill
    \label{img_models}
    
    \textbf{ResNet:} Residual Neural Networks (ResNet) \cite{ResnetArq}, introduced the idea of skip connections on CNNs. The ``identity shortcut connection" present in ResNets, usually skipping two or three layers in the model, allowing for efficient training of deep CNNs, which are known to suffer from the vanishing gradient problem. This problem is present in the back-propagation of the calculated gradients to earlier layers. As the gradient is propagated backwards, repeated multiplications might produce very small gradients, resulting in performance degradation. By having residual connections, residual nets learn residual functions concerning the layer inputs. Furthermore, instead of learning a direct mapping using stacked-layers, residual connections let these layers learn residual mappings. This means that instead of having a layer learning a desired mapping $H(x)$, regarding to the input $x$, residual connections allow to reframe this mapping as $F(x) := H(x) - x$, which can then be reframed to $H(x) := F(x)+x$. More, if the identity mapping is optimal, residuals can be pushed to zero, meaning that residual networks will, at least, have the same performance of networks using stacked-layers without residual connections.
    
    The ResNet architectures, with different depths, implemented were: ResNet18, ResNet34, ResNet50, ResNet101 and ResNet152, using the parameters defined in the original paper \cite{ResnetArq}.
    
    \textbf{DenseNet:} Densely Connected Convolutional Networks (DenseNet) \cite{huang2017densely} improved upon ResNet by proposing a type of CNN that utilizes dense connections between layers, using Dense Blocks. In this networks, every layer obtains additional inputs from all preceding layers, instead of traditional layers, in which their input is the output of the last layer or the output of the last layer plus a short connection. By allowing multiple parallel connections, DenseNets preserve the feed-forward network scheme, by having all layers obtaining concatenated outputs of all preceding layers, and passing its own feature-maps to subsequent layers. This allows every layer to receive a ``collective knowledge" from all past layers, mitigating the vanishing-gradient problem, encouraging feature reuse and allowing for a reduced number of parameters in each layer, since the feature maps continuously expand by concatenation with previous feature maps.
    
    DenseNet models achieved state-of-the-art results in many image tasks while requiring less computation and fewer model parameters than previous state-of-the-art CNNs. In this work, we have implemented DenseNet161 using the parameters detailed in the original paper \cite{huang2017densely}.
    
    For both ResNet and DenseNet models implemented, we have also applied transfer learning and conducted an experiment, in which the initial layer of each model was modified to have 4 input channels, instead of the original 3. A detailed explanation of these experiments is presented in Section \ref{exp:image}.

\subsection{Text Sentiment Analysis}
\label{text_analysis}
    This section presents the operations performed over textual data and the models used to do so. For every deep learning model with an Embedding Layer, a GloVe 200 dimension pre-trained on Twitter Embedding was used \cite{pennington2014glove}.

    \subsubsection{Pre-processing}\hfill
    \label{text_pre}

    To effectively classify text, before inputting the text into a classifier, a pre-processing step takes place. This is done because: 1) the type of text obtained from social-media varies substantially and contains noise: syntactic, semantic and grammar errors, mainly due to size constraints, typing speed and slang; 2) standardize data, so that classifiers can more easily learn patterns; 3) be in concordance with the input constraints of Word Embeddings layers and different classifiers.

    So, the steps to clean the textual data were: 1) transform HTML codes into words and symbols; 2) remove stop words using NLTK functionalities; 3) transform every word to lower case; 4) remove occurrences of more than three equal sequential characters into a maximum of two (e.g., \emph{``sooo happppyy"} becomes \emph{``so happy"}); 5) remove links and specific social media user-mentions (both the mention and the ``RT" word from Twitter); and finally, 6) punctuation was removed.

    \subsubsection{Models}\hfill
    \label{text_models}

    \textbf{VADER:} Valence Aware Dictionary for Sentiment Reasoning (VADER) \cite{hutto2014vader}, is a lexicon and rule-based method designed to classify the polarity of text from social media. It uses a list of lexical features, such as words, which are labelled accordingly to their polarity. The output of VADER is based on the probability of the sentence belonging to either the positive, negative or neutral class. This method was created having as basis a set of human-curated lexicon sentiment analysis.

    \textbf{TextBlob:} TextBlob is a library that implements methods for processing textual data~\cite{loria2018textblob}. Like VADER, TexBlob sentiment analysis is based on a set of lexicons that are labelled accordingly to their polarity. This method performs an average over all the lexicons that represent the words in the input sentence and outputs a polarity between $-1$ and $1$. Using this polarity, we defined that values under $-0.1$ are classified as having a negative meaning, over $0.1$ are classified as positive, and the remainder is classified as neutral.

    \textbf{FastText:} FastText is a shallow network architecture that can be trained in a reduced time, when compared to deeper architectures, whilst achieving competitive results in multiple text processing tasks \cite{joulin-etal-2017-bag}. The idea behind FastText is to have an architecture, based on the Continuous Bag of Words (CBOW) model \cite{DBLP:journals/corr/abs-1301-3781} to represent words, being then followed by linear classifiers that classify the input.

    The FastText architecture designed in this work consists on an Embedding layer followed by two Linear layers, the first having input size equal to the embedding size and output equal to 256, whilst the second one has input size of 256 and outputs the classification vector.

    \textbf{LSTM:} The Long-Short Term Memory architecture (LSTM) \cite{doi:10.1162/neco.1997.9.8.1735}, is a variation of the traditional Recurrent Neural Networks (RNNs), and was created to mitigate the exploding and vanishing gradient problems found in simple RNNs. An LSTM layer consists of a predefined set of recurrent blocks, each one of them containing cells. These cells have three types: the input, output and forget gates and serve the purpose of updating, or not, the cell state, erasing its memory and deciding if the cell output should be available. With this architecture, an LSTM layer can store information for later use, preventing gradients from vanishing during the learning process, and can also determine what information to ignore, therefore, allowing the network to remember important information for a longer period of time.

    Formally, an LSTM layer computes for each element in the input sequence:
    \begin{align}
        \begin{split}\label{eq:1}
            i_t ={}& \sigma(W_{ii} x_t + b_{ii} + W_{hi} h_{(t-1)} + b_{hi})
        \end{split} \\
        \begin{split}\label{eq:2}
            g_t ={}& \tanh(W_{ig} x_t + b_{ig} + W_{hg} h_{(t-1)} + b_{hg})
        \end{split} \\
        \begin{split}\label{eq:3}
            o_t ={}& \sigma(W_{io} x_t + b_{io} + W_{ho} h_{(t-1)} + b_{ho})
        \end{split} \\
        \begin{split}\label{eq:4}
            c_t ={}& f_t \circ c_{(t-1)} + i_t \circ g_t
        \end{split} \\
        \begin{split}\label{eq:5}
            h_t ={}& o_t \circ \tanh(c_t)
        \end{split}
    \end{align}
    where $h_t$ is the hidden state at time $t$; $c_t$ is the cell state at time $t$; $x_t$ is the input at time $t$; $h_{(t-1)}$ is the hidden state of the layer at time $t-1$ or the initial hidden state at time $0$, and $i_t$, $f_t$, $g_t$, $o_t$ are the input, forget, cell, and output gates, respectively. $\sigma$ is the sigmoid function, and $\circ$ is the element-wise product.

    The architecture implemented consist of an Embedding layer, followed by an LSTM layer with an input size equal to the dimension of the Embedding and with the number of features in the hidden state equal to 256. This is then followed by a Linear layer with input size equal to the number of features in the LSTM (256) and output size equal to the number of classes.

    \textbf{LSTM-Attn:} The second LSTM architecture we implemented, LSTM-Attn, to perform sentiment analysis in the text is based on the aforementioned LSTM architecture, but with the addition of an Attention layer \cite{bahdanau2014neural} between the LSTM layer and the Linear layer. The idea behind the Attention mechanism is to increase the importance of specific parts of the input sentence \cite{DBLP:conf/iclr/LinFSYXZB17}. There are two types of attention mechanisms, the global and the local ones. The difference is that in the global mechanisms, all the hidden states of the previous layer are considered for deriving the context vector, whilst on the local attention mechanisms, only some hidden states are considered \cite{luong-etal-2015-effective}. In this implementation, the Attention layer is a global attention mechanism that computes the soft alignment score between the output of the LSTM and its final hidden state.
    
    The final architecture for LSTM-Attn is: an Embedding layer, followed by an LSTM layer with input size equal to the embedding dimension and $256$ as the number of hidden features. Following this, comes the Attention layer that receives the output from the LSTM and the last hidden LSTM state, and outputs a new hidden state with the same size as the output of the LSTM. This then serves as input to the Linear layer, which outputs the classification vector.
    
    \textbf{Bi-LSTM:} Bi-directional LSTMs (Bi-LSTM) were created as an improvement over the vanilla LSTMs in tasks where full sequences are present, and context is important \cite{GRAVES2005602}. The idea behind Bi-LSTM is to present the information forward and backwards to two separate LSTM networks that are both connected to the same output layer. Having such bi-directional processing on the input information means that the network has, for each input (embedding or word), sequential and context information about it.
    
    The architecture implemented starts with an Embedding layer, followed by a bi-directional LSTM layer with input size equal to the embedding dimension and $256$ as hidden features. Then, we use the output from the Bi-LSTM layer and perform both an average pool and a max pool, which are concatenated together and fed into a Linear layer of input size $256*4$ and output of $64$. Then, a ReLU operation is performed, followed by a dropout, with $p=0.1$. The result of this goes to a Linear layer that outputs the classification vector.
    
    \textbf{RNN:} Recurrent Neural Networks were designed to allow neural networks to have temporal information, which simple neural networks cannot have \cite{elman1990finding}. Basically, RNNs form a chain structure in which each node receives as input the output from the predecessor node and one part of the input sequence (e.g., a word or a vector). Each node outputs a value, both to the successor node and to the next layer.
    
    So, what an RNN layer does is, for each input element, it computes:
    \begin{equation}
        h_t = \text{tanh}(W_{ih} x_t + b_{ih} + W_{hh} h_{(t-1)} + b_{hh})
    \end{equation}
    
    where $h_t$ is the hidden state at time $t$; $x_t$ is the input at time $t$, and $h_{(t-1)}$ is the hidden state of the previous layer at time $t-1$ or the initial hidden state at time $0$.
    
    The architecture implemented is an Embedding Layer followed by a multi-layer Elman RNN with $2$ layers, input size equal to the dimension of the embedding and with a hidden size of $256$. The output of this layer is then inserted into a Linear layer that outputs the classification vector.
    
    \textbf{RCNN:} The idea behind Recurrent Convolutional Neural Network (RCNN) \cite{Lai:2015:RCN:2886521.2886636}, is to apply a recurrent network structure to text classification that requires no human-designed features. The recurrent structure captures contextual information as far as possible when learning word representations whilst having less noise when compared to methods that use neural networks that rely on window-based processing. To store the context, the RCNN uses a bi-directional recurrent structure and employs a max-pooling layer to capture key features present in the text automatically.
    
    The architecture implemented consists of an Embedding Layer, a bi-directional LSTM Layer with input size equal to the dimension of the embedding, hidden size of $256$ and a dropout of $0.8$. The final embedding vector is the concatenation of its embedding and left and right contextual embeddings, which in this case is the hidden vector of the LSTM. This concatenated vector is then passed to a Linear Layer which maps the input vector back to a vector with a size equal to the hidden size of the LSTM, $256$. This is passed through a 1D Max Pooling Layer, and finally, the output from this layer is sent to a Linear Layer that maps the input to a classification vector.
    
    \textbf{TextCNN:} The Convolutional Neural Networks for Sentence Classification (TextCNN), performs convolutions, of different kernel sizes, on textual data \cite{kim-2014-convolutional}. This is done by performing convolutions over the embedding matrix that represents the input sentences. The convolutions performed are parallel and independent of one another. The architecture of TextCNN implemented is an Embedding Layer followed by $5$ convolutional blocks. Each one of those blocks consists of a 2D Convolutional Layer followed by a ReLU activation function and a 1D Max Pooling Layer. The output of all blocks is concatenated into a single vector that goes through a dropout phase with $p=0.8$, and the result then goes to a Linear Layer that returns the classification vector.
    
    Based on TextCNN, we defined a second CNN-based network to perform text classification, which we named \textbf{sCNN}. The architecture is similar, but instead of having $5$ blocks of Convolutions-ReLU-Max Pooling, it has only $3$ with kernel sizes of $1$, $3$ and $5$ respectively.

    \textbf{VDCNN:} Inspired by the results that deep convolutional networks have on image classification tasks \cite{simonyan2014very}, the authors of Very Deep Convolutional Networks for Text Classification (VDCNN) \cite{conneau-etal-2017-deep} designed a similar deep neural network for the problem of classifying text. In VDCNN, many convolutions with small kernel sizes (size $3$) are stacked to form a deep network, where shortcuts are present for keeping contextual information and solving the vanishing gradient problem. VDCNN employs convolutional blocks that consist of a sequence of two convolutional layers, each one followed by batch normalization and a ReLU activation.
    
    We implemented 4 VDCNN architectures with different depths: 9, 17, 29 and 49. Every architecture starts with an Embedding layer, followed by a 1D Conv layer with input size equal to embedding size and output size of $64$. Then, they have a set of Convolution Blocks. The number of Convolution Blocks depends on the depth of the architecture, but can be seen in \cite{conneau-etal-2017-deep}-Table 2. After the Convolution Blocks, comes a K-Max Pooling layer, a Linear layer with input of $512*k$, where $k$ is the number selected for the pooling layer, $(2)$, and output of $2048$. Following it, a Linear Layer with $2048$ as input and output is inserted and finally, a Linear Layer with an input size of $2048$, outputs the classification vector.

\subsection{Fusion Method}
\label{fusion:automl}
    The focus of the fusion method is to use the individual classifications of the text and image components to perform a final classification. The goal with the fusion approach is to leverage context knowledge of both sources, to outperform individual classifications. 
    
    To create the classifier, we based our approach on AutoML \cite{automlbook}, where the goal is to create an optimal model to classify a given dataset, without requiring extensive human modelling. First, we can define a machine learning model $\mathbb{L}$, as a mapping from the space of datasets, $D$, and architectures, $A$, to the space of models $M, \mathbb{L} : D \times A \to M$. For any given dataset $d \in D$, and architecture $a \in A$, the mapping returns the solution to the problem, which consists of minimizing a loss function, $\mathcal{L}$, with regularization mechanisms, $R$, with respect to the model, $m$, with parameters $\theta$, architecture $a$, and using the training data, $d^{(train)}$ \cite{martintransfer}:
    \begin{equation}
        \mathbb{L}(a,d^{(train)}) = \argminA_{m^{(a,\theta)} \in M^{(a)}} \mathcal{L}(m^{(a,\theta)}, d^{(train)}) + R(\theta)
    \end{equation}
    
    So, we can define our problem as a nested optimization problem, where the goal is to find an optimal model to classify the sentiment based on the individual classifications, $d$, and a search space $A$: $a^* \in A$, that maximizes the objective function $\mathcal{O}$, on the validation set:
    
    \begin{equation}
        a^* = \argmaxA_{a \in A} \mathcal{O}(\mathbb{L}(a,d^{(train)}), d^{(valid)})
    \end{equation}
    
    Since our problem relies on fusing individual classifications of both text and image into a final one, the first step is to get $d$ based on $Y_{img}$ and $Y_{text}$, where $Y$ represents the classification vector, and $img$ and $text$ represent the classifiers, in order to fuse them into a unique feature map: $X = Y_{img} \bigoplus Y_{text}$, where $X$ will be the input for the optimization problem (final classifier). Note that in our problem, $\mathcal{O}$ is defined as accuracy in the task of 3-class sentiment classification.

    To search for the optimal machine learning model, we based our solution on \cite{H2OAutoML}, by performing an automatic random search \cite{bergstra2012random} over a set of several machine learning algorithms and their inner parameters. So, in this work, to search the optimal model and its inner parameters, we performed a random search over the space that includes the following models: a random forest, an extremely-randomized forest, a random grid of generalized linear models, a random grid of XGboost, a random grid of gradient boosting machines (GBM), a random grid of deep neural networks. After searching these models, 2 stacked ensembles were created, the first one comprised of all models evaluated, and the other one, containing the best model of each type. In the end, the model with the best performance on the validation set is the one selected to be in the architecture of the proposed method.

\section{Experiments}
\label{experiments}

\subsection{Datasets}
    As aforementioned, our approach tackles the problem of multimodal sentiment analysis, using both textual and image information. For this, we focused on using the B-T4SA dataset, which is a dataset comprised of Twitter information, in which every sample has both text and image. Furthermore, to conduct experiments using transfer-learning \cite{pan2009survey}, we incorporated two more datasets: \textit{Stanford Sentiment Treebank} (SST-5), for the text classification, and \textit{Flickr and Instagram Dataset}, for the image classification component.
    
    Following is a detailed explanation of all the datasets used.

\subsubsection{B-T4SA}
\label{dataset:bt4sa}

\begin{figure*}[!t]
        \centering
        \subfigure[\textbf{Negative}: ``His eyes speak of the horror of war that no one should go through. War doesn't help, it only kills. Pls Stop.'']{\includegraphics[height=4cm,width=0.30\textwidth]{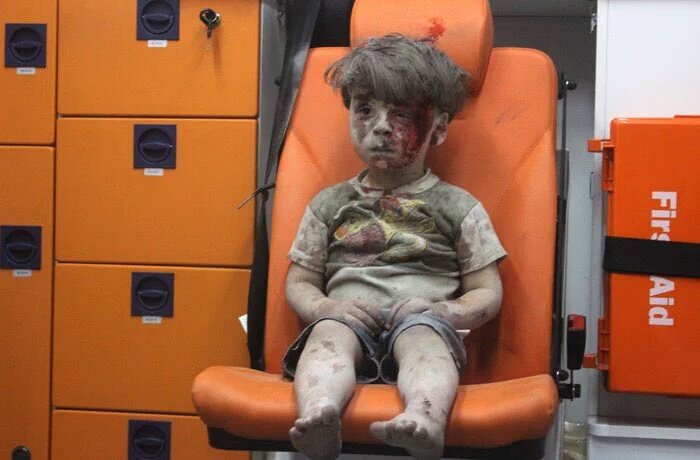}}
        \hspace{1em}
        \subfigure[\textbf{Neutral}: ``And they say it's grim up north...'']{\includegraphics[height=4cm,width=0.31\textwidth]{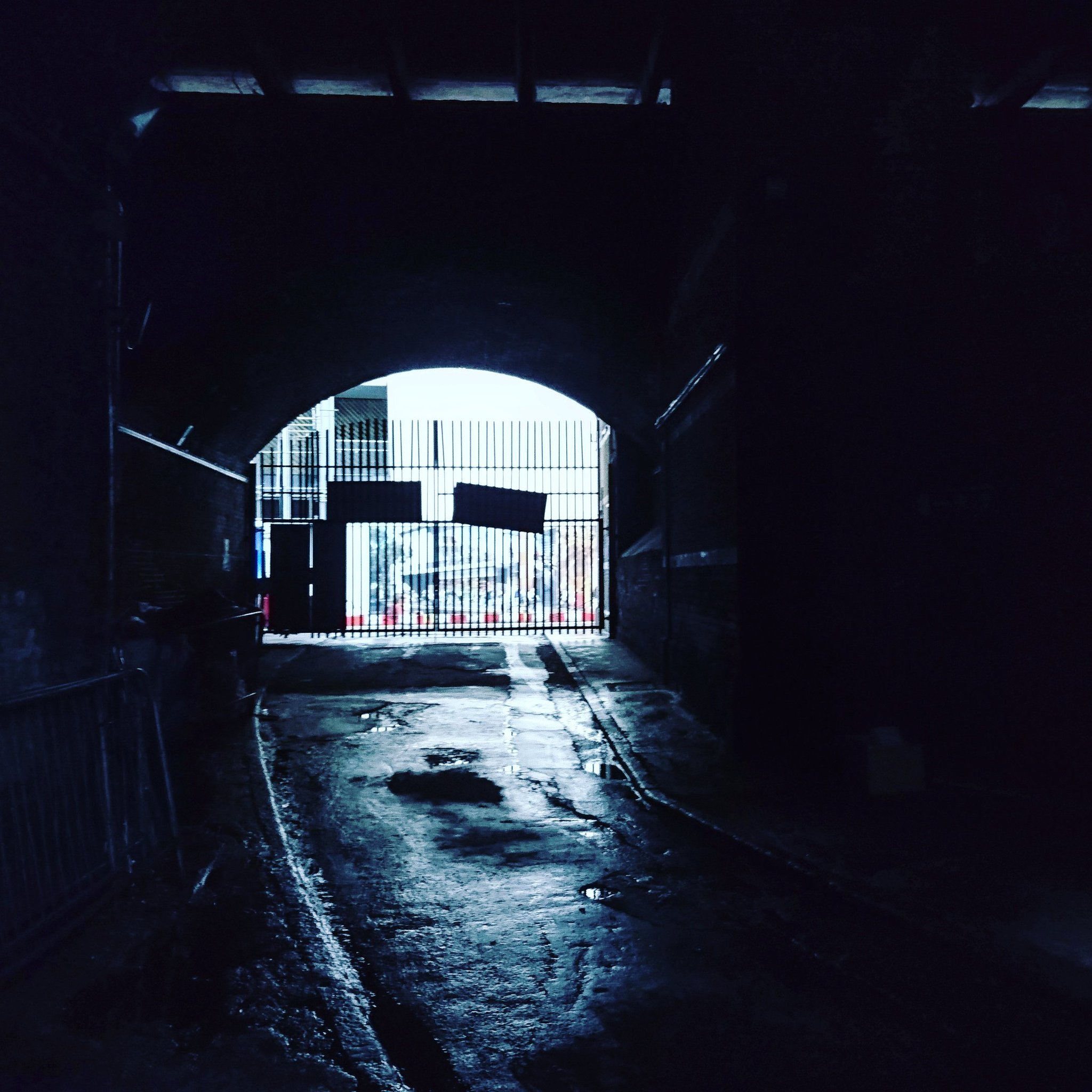}}
        \hspace{1em}
        \subfigure[\textbf{Positive}: ``Thank you Phuket Sunset Weddings for using Wedding Flowers Phuket. Waiting the happy couple.'']{\includegraphics[height=4cm,width=0.31\textwidth]{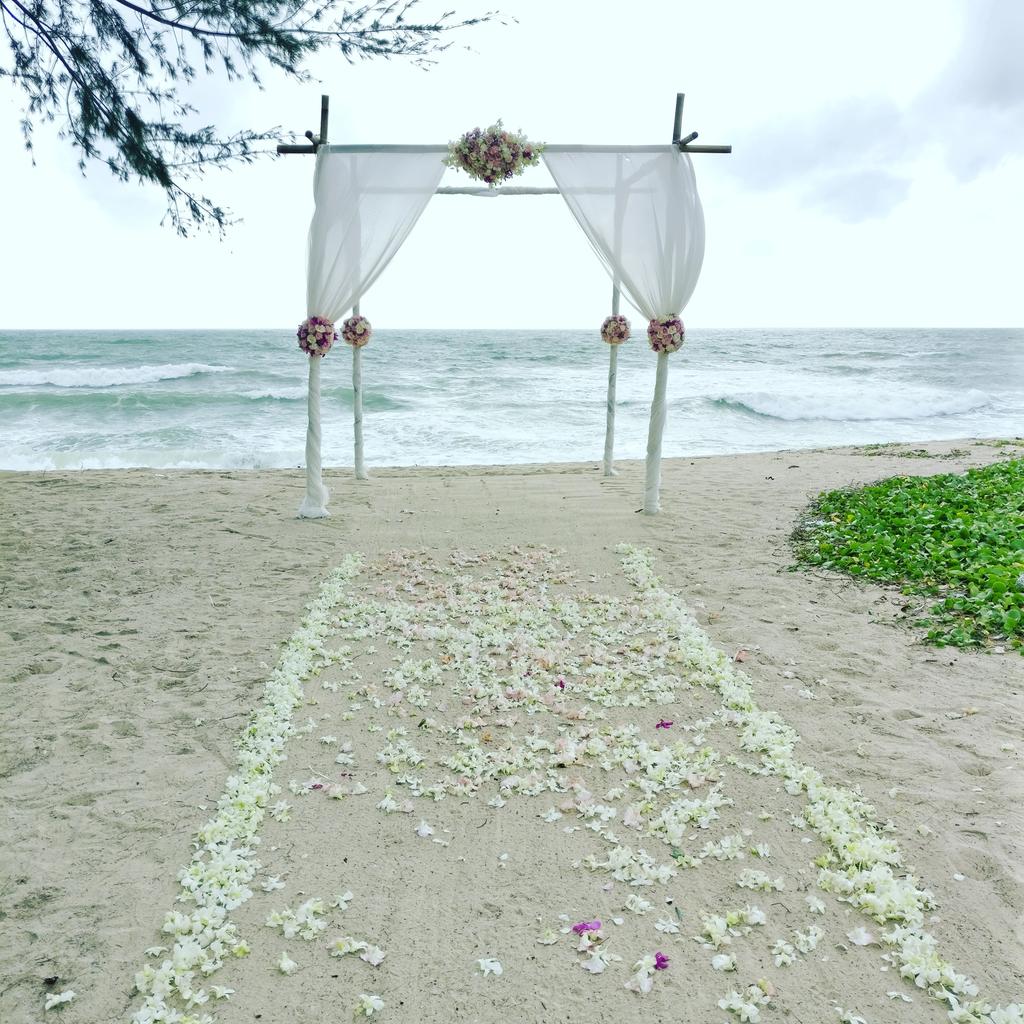}}
    
        {\caption{Examples of images and the correspondent texts of the three different classes presented in the dataset. (a), presents a negative example; (b), a neutral one, and (c), a positive sample. \label{figdataset}}}
    \end{figure*}

    B-T4SA is a subset of T4SA, consisting of 470 thousand samples, each one containing both text and image information. All classes are balanced, and the splits are stratified. The train set consists of approximately 80\% of the dataset, while both the validation and test sets have 10\% each. B-T4SA was created to solve the problems of T4SA, such as duplicated entries, small sentences, malformed images, and unbalanced classes \cite{8265255}. In Figure \ref{figdataset}, we show an example of an image and the corresponding text, for each class (negative, neutral, positive).
    
\subsubsection{Stanford Sentiment Treebank}
    The Stanford Sentiment Treebank (SST) consists of sentiment labels for 215,154 phrases in the parse trees of 11855 sentences \cite{socher2013recursive}. The dataset can be presented in the form of binary classification, either negative or positive, or in a fine-grained way, using a 5-class classification: very negative, negative, neutral, positive, and very positive. The latter is denominated SST-5, and is widely used to evaluate text sentiment classifiers. SST-5 was used in this work as a way to initially train a model, before training it on a final dataset. This allows performing fine-tuning on the model, by transferring the knowledge from the first dataset to the second one, avoiding initializing the model weights randomly \cite{weiss2016survey}. Denote that, as this dataset has 5 classes, when the models were transferred to B-T4SA, the last layer was removed and substituted by a new one with only 3 output classes.
    
\subsubsection{Flickr and Instagram}
    To perform transfer-learning on the image classifier, we have also used the Flickr and Instagram dataset \cite{10.5555/3015812.3015857}, which is composed of 23308 labelled images of 8 different classes of emotions - amusement, anger, awe, contentment, disgust, excitement, fear and sadness. The goal with this dataset was to pre-train the image classifiers, which upon convergence, are transferred to B-T4SA, by replacing the last classification layer to an identical one with only 3 output classes.

\subsection{Text Analysis}
    To select the best text sentiment analysis model to use in the multimodal architecture, we conducted a set of experiments with all the models implemented. We evaluated each model three times in the task of classifying sentiments in the B-T4SA dataset, using the Adam optimizer \cite{DBLP:journals/corr/KingmaB14}, and the Cross-Entropy loss.

    The mean accuracy and standard deviation in each one of the sets are shown in Table \ref{tab:textanalysis}, where the first block represents the results for two models that can be used as a library in python (VADER and TextBlob), the second block represents the deep learning models, and the third block represents the best model of the previous block with fine-tuning. In the first block, we show two results for both methods, including with and without the pre-processing step. The ``-PP" represents the results with clean data, which achieved better results than without any data cleaning, showing that cleaning textual data to remove noise will allow methods to yield better results. In the second and third blocks, all experiments were conducted using pre-processed data. Here, it is possible to see that, except for FastText that achieved approximately 42\% and LSTM that was incapable of learning to solve the task (even with different learning rates, hidden features and optimizers), all models achieved a mean accuracy of over 90\%. This is well above the plug-and-play methods, and the previous state-of-the-art \cite{AGasparIdeal}, which used TextBlob methods to achieve 64.27\% accuracy. The best result from these methods was obtained with the RCNN, which achieved a mean accuracy of 94.61\%, outperforming all other methods. This can be justified due to the strong capability of RCNNs to evaluate a word, based on its embeddings, coupled by the right and left contexts, which are extracted using recurrent structures. This combination allows features to be extracted more accurately when working in problems that require context, of which sentiment analysis is heavily dependent on. On the third block of the table, the results of fine-tuning the RCNN model are presented. In this case, the fine-tuning was performed by initially training the model on the SST-5 dataset, and then replacing its final classification layer to output three values instead of five. More, ``RCNN-sst \textit{ft} B-T4SA FC" represents training initially on SST-5 and then train only on the Linear Layers of the model using B-T4SA, whereas ``RCNN-sst \textit{ft} B-T4SA" represents fine-tuning the entire model on B-T4SA, after training it on SST-5. By conducting such experiments, it is possible to see that fine-tuning the entire model yields better results when compared to only transfer learning and fine-tune the last classification layers. However, these results do not improve upon the normal model, with weights initialized randomly.

    From this experiment, we selected RCNN as the text classifier to be used in the proposed method, since it presented the best results in the validation set. Even though the validation set cannot be seen as a surrogate of the test set, it is the best way to evaluate how a model will perform in unseen data, without introducing biases by using the test set, which is only used at the end of the entire training process (when all the methods for the proposed method are selected).

\begin{table}[]
    \caption{Mean accuracy and standard deviation on the train and validation set of the B-T4SA dataset with different methods for language processing. The first block shows the results using methods that are available as python libraries. The second block shows the result of our implementations of different deep learning methods. In the third block, we show the results of the best method from the second block (RCNN), pre-training on SST and fine-tuned on B-T4SA. Each model was evaluated three times, under the same conditions.}
    \label{tab:textanalysis}
    \centering
        \begin{tabular}{p{3cm}|lll} 
                                                 & \multicolumn{2}{c}{\textbf{Mean Accuracy (\%)}}                                                              \\
            \multicolumn{1}{c|}{\textbf{Method}} & \multicolumn{1}{c}{Train}                       & \multicolumn{1}{c}{Validation} \\ \hline 
            VADER                                & 41.04 $\pm~0$                                   & 41.02 $\pm~0$                  \\ 
            VADER-PP                             & 56.84 $\pm~0$                                   & 56.82 $\pm~0$                  \\ 
            Textblob                             & 64.22 $\pm~0$                                   & 64.27 $\pm~0$                  \\ 
            Textblob-PP                          & 64.88 $\pm~0$                                   & 64.78 $\pm~0$                  \\            
            \hhline{----}
            FastText                             & 42.86 $\pm~0.03$                                & 42.76 $\pm~0.05$               \\ 
            LSTM                                 & 33.33 $\pm~0.04$                                & 33.13 $\pm~0.00$               \\ 
            LSTM-Attn                            & 97.36 $\pm~0.03$                                & 93.48 $\pm~0.57$               \\
            BI-LSTM                              & 96.56 $\pm~0.97$                                & 94.35 $\pm~0.07$               \\ 
            RNN                                  & 90.72 $\pm~0.78$                                & 91.24 $\pm~0.48$               \\ 
            RCNN                                 & 98.12 $\pm~1.10$                                & \textbf{94.61 $\pm~0.03$}      \\ 
            TextCNN                              & 95.47 $\pm~0.86$                                & 93.73 $\pm~0.00$               \\ 
            sCNN                                 & 90.69 $\pm~0.10$                                & 92.69 $\pm~0.02$               \\ 
            VDCNN9                               & 94.18 $\pm~0.81$                                & 93.33 $\pm~0.23$               \\ 
            VDCNN17                              & 88.29 $\pm~2.28$                                & 92.05 $\pm~0.52$               \\ 
            VDCNN29                              & 93.90 $\pm~0.65$                                & 93.19 $\pm~0.18$               \\ 
            VDCNN49                              & 92.61 $\pm~0.33$                                & 92.81 $\pm~0.11$               \\ 
            \hhline{----}
            RCNN-sst \textit{ft} B-T4SA FC       & 86.73 $\pm~0.69$                                & 86.51 $\pm~0.67$               \\

            RCNN-sst \textit{ft} B-T4SA          & \textbf{98.60 $\pm~1.29$ }                      & 94.60 $\pm~0.03$                 
        \end{tabular}%

\end{table}

\subsection{Image Analysis}
\label{exp:image}
    Regarding the selection of the model to perform the image classification, we have evaluated the performance of multiple resnet architectures and densenet161 in the task of image sentiment analysis. In Table~\ref{tab:netresults}, the results for our experiments are shown. Note that every network was evaluated using the same learning rate ($1e-3$), Adam optimizer and cross-entropy loss. The first row of the table represents if the experiment was done using transfer-learning, meaning that the models were initially trained on the Flicker and Instagram dataset. We have also evaluated how the different models behave with RBG images (1st and 4th experiment in the table, while the 4th uses pre-trained weights), and RBG and Local Binary Patterns (LBP) \cite{5972286}. In the latter, we changed the models to receive 4 inputs and placed the LBP on the fourth channel (3rd experiment in the table). The goal of using LBP is to evidence hidden patterns that assist in detecting the image polarity. All the results show that classifying the sentiment of an image is difficult, mainly due to the subjectivity of the image and due to inter-class similarities, where images that have different classes can be visually similar. Neither the addition of the LBP nor pre-training the models on the Flicker and Instagram dataset improved the results when compared to using only RGB with randomly initialized weights. Furthermore, all models performed similarly, but ResNet34 was the best one, achieving 49.8\% accuracy using RGB images, with or without pre-training. Even though ResNet18 had almost the same performance using pre-trained settings, we selected ResNet34 for the proposed method, as it consistently out-performed ResNet18.
    
    \begin{table}[!t]
    \caption{Accuracy (\%) of several deep learning networks in the task of classifying sentiments, both in the train and validation set. The first row, pre-train, indicates if the experiment uses transfer learning. Each experiment column also contains the information about the data type used, either RGB or RGB plus LBP.}
    \label{tab:netresults}
    \begin{center}
        \begin{tabular}{c c c| c c| c c }
            \hline
            \textbf{Pre-trained} & \multicolumn{2}{c}{$\times$}            & \multicolumn{2}{c}{$\times$}       & \multicolumn{2}{c}{\checkmark}      \\
            \hline
          
            \multicolumn{1}{c}{\multirow{2}{*}{\textbf{Network}}}                   & \multicolumn{2}{c|}{\textbf{RGB}}                                          &                            \multicolumn{2}{c|}{\textbf{RGB+LBP}}                                              & \multicolumn{2}{c}{\textbf{RGB}}                                                                                                                                                                                                                                                                                                 \\
                                                                  & \multicolumn{1}{c}{\textbf{Train}} & \multicolumn{1}{c|}{\textbf{Val}}& \multicolumn{1}{c}{\textbf{Train}} & \multicolumn{1}{c|}{\textbf{Val}}  & \multicolumn{1}{c}{\textbf{Train}} & \multicolumn{1}{c}{\textbf{Val}}   \\
            \hline
            \textbf{ResNet18}    & \textbf{47.4}\% & 47.7\%                    & 47.4\%          & 47.9\%          & 46.6\%          & 49.7\%  \\
            \textbf{ResNet34}    & 47.2\%          & \textbf{49.8}\%           & 47.3\%          & \textbf{48.0}\% & 45.6\%          & \textbf{49.8}\%  \\
            \textbf{ResNet50}    & 46.3\%          & 46.4\%                    & 47.2\%          & 47.4\%          & \textbf{48.5}\% & 48.7\%  \\
            \textbf{ResNet101}   & 44.9\%          & 45.1\%                    & 47.1\%          & 47.1\%          & 47.6\%          & 47.7\%  \\
            \textbf{ResNet152}   & 44.5\%          & 44.5\%                    & 45.9\%          & 45.9\%          & 47.1\%          & 47.5\%  \\
            \textbf{DenseNet161} & 46.9\%          & 47.1\%                    & \textbf{47.5}\% & 47.5\%          & 47.2\%          & 47.3\%  \\

            \hline
        \end{tabular}
    \end{center}
\end{table}

\subsection{Fusion Analysis}
    Based on the text classifier, RCNN, and the image classifier, ResNet34, we then evaluated the performance of the proposed method as a whole. For this, we initially searched for the optimal model, using the method described in \ref{fusion:automl}, allowing the search for a maximum of two hours. By doing this, the most performant model in the validation set was a GBM. By evaluating the entire proposed method on the test set, it achieved an accuracy of 95.19\%, which the result is synthesized in Table \ref{tab:fusion_methods}. To evaluate the performance of the proposed method, but with a different fusion classifier, we have also evaluated the use of a Support Vector Machine (SVM), which achieved a performance of 95.16\% using all the settings of the proposed method. The difference between the GBM and the SVM classifier is small, 0.03\%, but the AutoML searched method has several advantages. The first one is the time required to train, while the AutoML method of searching for methods only required two hours, SVM required several hours to train. More, SVMs tend not to scale well, as there are more features (in the order on thousands), they tend to become extremely slow to fit the data, whilst our proposed methodology to search for a classifier is extremely robust by comprising methods that can handle a large number of features without becoming untenable. However, the SVM baseline consolidates the proposed method, by showing that the proposed architecture works, even in the presence of different fusion classifiers.
    
    To further validate our proposal, we compare our results with state-of-the-art methods, in which the results are present in Table \ref{tab:finalresults}. In this, it is possible to see that our proposal outperforms others. More, to further evaluate the effectiveness of our proposal, we have further tested the proposed method in \cite{AGasparIdeal}, by replacing its text classifier by our RCNN, resulting in a 15.9\% accuracy improvement, (represented in the table by Information Fusion \cite{AGasparIdeal} (TM)), but is still 18.8\% below our proposed method. This consolidates that our proposed method of fusing the individual classifications and then performing a random search to find the optimal fusion classifier, is an efficient method.
    
    \begin{table}[!t]
    \centering
    \caption{Accuracy (\%) in the test set for the proposed method and a baseline using SVM.} 
    \label{tab:fusion_methods}
    {
        \setlength{\tabcolsep}{8pt}
        \begin{tabular}{c c c c}
            \hline
            \textbf{Method}  & \textbf{Test Accuracy (\%)} \\ 
            \hline
            SVM                     & 95.16\%           \\ 
            AutoML-based Fusion \textbf{(ours)}              & \textbf{95.19\%}  \\ 
            \hline
        \end{tabular}
    }
\end{table}

\begin{table}[!t]
    \caption{Comparison with the results of existing methods with our experiments. The TM, stands for substituting the text classifier from \cite{AGasparIdeal} for the one selected in our experiments.}
    \label{tab:finalresults}
    \centering

    \begin{tabular}{l|c}
        \multicolumn{1}{c|}{\textbf{Method}}   & \begin{tabular}[c]{@{}c@{}}\textbf{B-T4SA}\\ \textbf{Test Set Accuracy (\%)}\end{tabular} \\ \hline
        Random Classifier                      & 33.33\%                    \\
        Hybrid-T4SA FT-F \cite{8265255}        & 49.90\%                    \\
        Hybrid-T4SA FT-A \cite{8265255}        & 49.10\%                    \\
        VGG-T4SA FT-F \cite{8265255}           & 50.60\%                    \\
        VGG-T4SA FT-A \cite{8265255}           & 51.30\%                    \\
        Information Fusion \cite{AGasparIdeal} & 60.42\%                    \\
        Information Fusion \cite{AGasparIdeal} (TM) & 76.35\%                    \\
        SVM-fusion (\textbf{ours})                                & 95.16\%                    \\
        AutoML-based Fusion \textbf{(ours)}                             & \textbf{95.19\%}           \\
    \end{tabular}

\end{table}


\section{Conclusions}
\label{conclusions}
    This paper proposes a novel method to perform multimodal sentiment classification of social media content. The proposed method consists of performing individual text and image classifications, which are then fused by an AutoML-generated model to perform a final classification. We explored several state-of-the-art classifiers for both text and image. More, with the proposed AutoML approach, our method was capable of finding an optimal model that outperformed the state-of-the-art in the B-T4SA dataset, which, due to its natural content, is very challenging and contains intra and inter-class subjectivity.

\section*{Acknowledgments}
This work was supported by `FCT - Fundação para a Ciência e Tecnologia' throught the research grant `2020.04588.BD', partially supported by NOVA LINCS under grant UID/EEA/50008/2019, by the project MOVES-Monitoring Virtual Crowds in Smart Cities (PTDC/EEI-AUT/28918/2017) financed by FCT-Fundação para a Ciência e a Tecnologia, and partially supported by project 026653 (POCI-01-0247-FEDER-026653) INDTECH 4.0 – New technologies for smart manufacturing, cofinanced by the Portugal 2020 Program (PT 2020), Compete 2020 Program and the European Union through the European Regional Development Fund (ERDF).

\bibliographystyle{IEEEtran}
\bibliography{references}

\end{document}